\documentclass{article}

\usepackage{arxiv}
\usepackage{multirow,graphicx,enumerate,color,amsfonts,amsthm,amssymb,mathrsfs,dsfont,textcomp,subfigure,array,float,enumitem,xcolor,cancel,soul}

\usepackage[utf8]{inputenc} 
\usepackage[T1]{fontenc}    
\usepackage{hyperref}       
\usepackage{url}            
\usepackage{booktabs}       
\usepackage{amsfonts}       
\usepackage{nicefrac}       
\usepackage{microtype}      
\usepackage{lipsum}		
\usepackage{graphicx}
\usepackage{amsmath}

\title{Modified Possibilistic Fuzzy C-Means Algorithm for Clustering Incomplete Data Sets}

\author{ \href{http://orcid.org/0000-0001-8331-5793}{\includegraphics[scale=0.06]{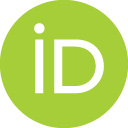}\hspace{1mm}Rustam}\thanks{Jl. Telekomunikasi No.1 Dayeuh Kolot, Kabupaten Bandung, Jawa Barat, Indonesia 40257 \url{(https://rustamtelu.staff.telkomuniversity.ac.id/)}} \\
Departement of Telecommunication Engineering\\ 
Faculty of Electrical Engineering\\ 
Telkom University\\
\texttt{rustamtelu@telkomuniversity.ac.id} \\
\And
\href{http://orcid.org/0000-0002-5228-1348}{\includegraphics[scale=0.06]{orcid.png}\hspace{1mm}Koredianto Usman} \\
Departement of Telecommunication Engineering\\ 
Faculty of Electrical Engineering\\ 
Telkom University\\
\And
\href{http://orcid.org/0000-0001-6932-1150}{\includegraphics[scale=0.06]{orcid.png}\hspace{1mm}Mudyawati Kamaruddin} \\
AKBID Tahirah Al Baeti Bulukumba\\Sulawesi Selatan, Indonesia\\
\texttt{mudya07@gmail.com}
\And
\href{http://orcid.org/0000-0001-9353-456X}{\includegraphics[scale=0.06]{orcid.png}\hspace{1mm}Dina Chamidah} \\
Department of Biology Education\\ 
Faculty of Language and Science\\
Universitas Wijaya Kusuma Surabaya\\
Surabaya, Jawa Timur, Indonesia\\
\And
\href{http://orcid.org/0000-0001-9641-677X}{\includegraphics[scale=0.06]{orcid.png}\hspace{1mm}Nopendri} \\
Departement of Industrial Engineering\\ 
Faculty of Industrial Engineering\\ 
Telkom University\\
\And
\href{http://orcid.org/0000-0002-2688-070X}{\includegraphics[scale=0.06]{orcid.png}\hspace{1mm}Khaerudin Saleh} \\
Departement of Telecommunication Engineering\\ 
Faculty of Electrical Engineering\\ 
Telkom University\\
\And
\href{http://orcid.org/0000-0002-7698-1445}{\includegraphics[scale=0.06]{orcid.png}\hspace{1mm}Yulinda Eliskar} \\
Departement of Telecommunication Engineering\\ 
Faculty of Electrical Engineering\\ 
Telkom University\\
\And
\href{http://orcid.org/0000-0003-3316-0484}{\includegraphics[scale=0.06]{orcid.png}\hspace{1mm}Ismail Marzuki} \\
Department of Chemical Engineering\\
Universitas Fajar\\
Panakkukang, Makassar, Sulawesi Selatan, Indonesia\\
}

\hypersetup{
pdftitle={A template for the arxiv style},
pdfsubject={q-bio.NC, q-bio.QM},
pdfauthor={David S.~Hippocampus, Elias D.~Striatum},
pdfkeywords={First keyword, Second keyword, More},
}

\begin{document}
\maketitle

\begin{abstract}
Possibilistic fuzzy c-means (PFCM) algorithm is a reliable algorithm has been proposed to deal the weakness of two popular algorithms for clustering, fuzzy c-means (FCM) and possibilistic c-means (PCM). PFCM algorithm deals with the weaknesses of FCM in handling noise sensitivity and the weaknesses of PCM in the case of coincidence clusters. However, the PFCM algorithm can be only applied to cluster complete data sets. Therefore, in this study, we propose a modification of the PFCM algorithm that can be applied to incomplete data sets clustering. We modified the PFCM algorithm to OCSPFCM and NPSPFCM algorithms and measured performance on three things: 1) accuracy percentage, 2) a number of iterations to termination, and 3) centroid errors. Based on the results that both algorithms have the potential for clustering incomplete data sets. However, the performance of the NPSPFCM algorithm is better than the OCSPFCM algorithm for clustering incomplete data sets.
\end{abstract}


\section{Introduction}
Incomplete data sets are common in the real world. Incomplete data sets can be caused by failures during data collection, the problem of merging data from various data sources, data cleaning or data transfer problems \cite{himmelspach}. The main problem faced when trying to cluster incomplete data sets is the existing clustering algorithm cannot be used for clustering incomplete data sets. Popular clustering algorithms included \textit{fuzzy c-means} (FCM) \cite{bezdek1} and \textit{possibilistic c-means} (PCM) \cite{krishnapuram}, which can only be used for complete data sets. Bezdek and Hathaway \cite{bezdek2} have developed FCM algorithm to deal with the problem of clustering data sets contain missing values. They proposed \textit{whole data strategy fuzzy c-means} (WDSFCM) to deal with incomplete data sets clustering problems by removing features that contain missing values and running standard FCM algorithms on the remaining data that have become complete data sets. However, WDSFCM produces biased clustering results when the missing values are large. 

Dixon \cite{dixon} proposed \textit{partial distance strategy} (PDS) algorithm to deal with incomplete data sets in clustering by calculates a partial distance (squared euclidean) using all the values available at the data points containing missing values and then scaling this quantity by the reciprocal of the component proportion used. Bezdek and Hathaway \cite{bezdek2} modified FCM using PDS to deal with clustering incomplete data sets known as the PDSFCM algorithm. WDSFCM and PDSFCM algorithms do not impute missing values or in other words, do not get estimates of missing values after the clustering process. 

The following two algorithms, which were also proposed by Bezdek and Hathaway in \cite{bezdek2} have imputed missing values. They modified the FCM algorithm using the \textit{optimal completion strategy} (OCS) and the \textit{nearest prototype strategy} (NPS), each of which is referred to as the OCSFCM and NPSFCM algorithms. The \textit{optimal completion strategy fuzzy c-means} (OCSFCM) algorithm estimates missing values by considering missing values as an additional variable and partitioning the data together with optimizing the value of the FCM objective function. The \textit{nearest prototype strategy fuzzy c-means} (NPSFCM) algorithm estimates missing values using the prototype cluster closest to itself in each iteration step. So the difference between OCSFCM and NPSFCM algorithms lies in how to update the imputation for missing values at each iteration step.

In another paper, Bezdek et al. \cite{bezdek3} also introduced the \textit{possibilistic fuzzy c-means} (PFCM) algorithm. The PFCM algorithm corrects the shortcomings of the FCM and PCM algorithms by overcoming noise sensitivity and the possibility of the occurrence of coincidental clusters, respectively. However, the PFCM algorithm also has disadvantages like those of FCM and PCM algorithms, which can be only used for clustering complete data sets. Therefore, in this study we propose a modification of PFCM algorithm so that it can be applied to incomplete data sets. Incomplete data sets are overcome by a strategy adapted from Bezdek and Hathaway\cite{bezdek2}, using the \textit{optimal completion strategy} (OCS) and the \textit{nearest prototype strategy} (NPS). This modification of the PFCM algorithm uses OCS as the OCSPFCM algorithm. While the modification of the PFCM algorithm uses the NPS, we term as the NPSPFCM algorithm.

This paper will describe the PFCM algorithm for clustering complete data sets in Section \ref{section:2}. In Section \ref{section:3}, a modification of the PFCM algorithm for clustering incomplete data sets is explained. In Section \ref{section:4}, we describe the experimental setup. In Section \ref{section:5}, experimental results on real world and artificial data sets are shown and we discuss the results, and Section \ref{section:6} is the conclusion of this study.

\section{Possibilistic Fuzzy C-Means (PFCM) Algorithm of Complete Data Sets}\label{section:2}
Suppose unlabeled data sets $X=\left \{\textbf{x}_{1},\textbf{x}_{2},\cdots,\textbf{x}_{n} \right\}\subset \mathbb{R}^{p} \: (p = n \times s)$ will be clustered into a fuzzy subset of $c$ ($1<c<n$) clusters. Here $n$ states the number of data points and $s$ states dimension each of data point. The purpose of clustering $X$ into $c$ clusters is achieved by minimizing the following objective functions \cite{bezdek3}.
\begin{equation}
J_{m,\tau}(U,T,\textbf{V};X)=\sum_{k=1}^{n}\sum_{i=1}^{c}(\alpha u_{ik}^{m}+\beta t_{ik}^{\tau})d_{ik}^{2}+\sum_{i=1}^{c}\delta_{i} \sum_{k=1}^{n}(1-t_{ik})^{\tau}.
\label{eq:1}
\end{equation}

Here $\alpha(\alpha>0)$ states the importance level of fuzzy membership degree $(u_{ik})$. Equation (\ref{eq:1}) is subject to $\sum_{i=1}^{c}u_{ik}=1$ constraints. Krishnapuram and Keller \cite{krishnapuram} relaxed this constrain become $\sum_{i=1}^{c}u_{ik}\geq1$, so that it would be better in reflecting the typical of $\textbf{x}_{k}$ to the $i$-th cluster. $t_{ik}$ states possibilistic membership degree of $\textbf{x}_{k}$ to the $i$-th cluster. So $\beta(\beta>0)$ states the importance level of possibilistic membership degree ($t_{ik}$). $d_{ik}^{2}=\left \| \textbf{x}_{k} - \textbf{v}_{i}\right \|$ states the Euclidean distance of the $j$-th data to $i$-th cluster centre vector. $\textbf{V}=(\textbf{v}_{1},\textbf{v}_{2},\cdots,\textbf{v}_{c})$ states cluster centre vector, $\textbf{v}_{i}\in\mathbb{R}^{s}$ and $\delta_{i}>0$ is the typical of possibilistic. Here $m>1$ and $\tau>1$ are weighting exponent.

Basically determining of the $u_{ik}$, $t_{ik}$ and $\textbf{v}_{i}$  must be done simultaneously. However, in this study we determined numerically using the recursive method. So that we can choose which values will be initiated to calculate the other values. Here, we chose initiate $\textbf{v}_{i}$ to calculate $u_{ik}$ and $t_{ik}$.

\subsection{Possibilistic Fuzzy C-Means (PFCM) Algorithm}

In this study, the complete data sets is clustered use the \textit{possibilistic fuzzy c-means} (PFCM) algorithm proposed by Bezdek et al. \cite{bezdek3}. The PFCM algorithm is described as follows.

\textbf{Step I}: Fix $m>1$, $\tau>1$, $\epsilon > 0$ and $1<c<n$. Pick $\textbf{v}^{(0)} \in\mathbb{R}^{s}$, $\textbf{v}^{(0)}$ can be chosen randomly from $X=\left \{\textbf{x}_{1},\textbf{x}_{2},\cdots,\textbf{x}_{n} \right\}\in\mathbb{R}^{p}$. Then at step $l$, $l=1,2,\cdots$

\textbf{Step II}: Calculate fuzzy membership degree ($u_{ik}$) which minimize the objective function $J_{m,\tau}$ using the following
\begin{equation}
u_{ik}^{(l)}=\left(\sum_{j=1}^{c}\left(\frac{d_{ik}^{2}}{d_{jk}^{2}}\right)^{\frac{1}{m-1}}\right)^{-1}, 1 \leq  i \leq c ;~  1 \leq  k \leq  n.
\end{equation}

\textbf{Step III}: Calculate possibilistic typical ($\delta_{i}$) which minimizes the objective function $J_{m,\tau}$ using the following

\begin{equation}
\delta_{i}^{(l)}=\frac{\sum_{k=1}^{n}\left(u_{ik}^{(l)}\right)^{m}d_{ik}^{2}}{\sum_{k=1}^{n}\left(u_{ik}^{(l)}\right)^{m}},1 \leq  k \leq n.
\end{equation}

\textbf{Step IV}: Calculate possibilistic membership degree ($t_{ik}$) which minimizes the objective function $J_{m,\tau}$ using the following

\begin{equation}
t_{ik}^{(l)}=\left(1+\left( \frac{\beta}{\delta_{i}}d_{ik}^{2}\right)^{\frac{1}{\tau-1}}\right)^{-1}, 1 \leq i \leq c ;~ 1 \leq k \leq n.
\end{equation}

\textbf{Step V}: Update cluster centre ($\textbf{v}_{i}$) which minimizes the objective function $J_{m,\tau}$ using the following

\begin{equation}
\textbf{v}_{i}^{(l)}=\frac{\sum_{k=1}^{n} \left(\left(\alpha u_{ik}^{(l)}\right)^{m} + \left(\beta t_{ik}^{(l)}\right)^{\tau}\right) \textbf{x}_{k}}{\sum_{k=1}^{n} \left(\left(\alpha u_{ik}^{(l)}\right)^{m} + \left(\beta t_{ik}^{(l)}\right)^{\tau}\right)}, 1 \leq k \leq n.
\end{equation}

\textbf{Step VI}: Compare $\textbf{v}_{i}^{(l)}$ to $\textbf{v}_{i}^{(l-1)}$ using $\left\| \textbf{v}_{i}^{(l)}-\textbf{v}_{i}^{(l-1)}\right\|<\epsilon$. If true, then stop. Otherwise, set $l=l+1$ and return to \textbf{Step II}. 

The clustering result of the complete data set will be a base for evaluating the performance of the modified PFCM algorithm (see Section \ref{section:4}). 

\section{Possibilistic Fuzzy C-Means (PFCM) Algorithm of Incomplete Data Sets}\label{section:3}

As explained earlier, PFCM algorithm cannot be used to cluster incomplete data sets or data sets that contain missing values. Therefore, in this section the PFCM is modified so that it can be used to cluster incomplete data set. Given incomplete data sets $Y=\left \{\textbf{y}_{1},\textbf{y}_{2},\cdots,\textbf{y}_{n} \right\}\subset \mathbb{R}^{p} \: (p = n \times s)$, with $\textbf{y}_{2}=\left(2.35,?,1.32,?,3.44\right)^{T}\in \mathbb{R}^{5}$. $y_{22}$ and $y_{24}$ are missing values. The question is how to cluster $Y$? Therefore, a modification of the PFCM algorithm is proposed to clustering data sets similar like $Y$. The notation that will be used throughout follows. Let $\textbf{y}=k^{th}$ $s$-dimensional data vector, for $1 \leq k \leq n$; $y_{kj}=j^{th}$ feature value of the $k^{th}$ data point vector, for $1 \leq j \leq s, 1 \leq k \leq n$; $Y=\left\{\textbf{y}_{1},\textbf{y}_{2},\cdots,\textbf{y}_{n} \right\}\subset\mathbb{R}^{p}$; $Y_{C}=\left \{\textbf{y}_{C} \in Y| \textbf{y}_{C}\:is\:a \:complete \:data \: point \right\}$; $Y_{M}=\left \{\textbf{y}_{M} \in Y| \textbf{y}_{M}\:is\:a \:incomplete \:data \: point \right\}$.

\subsection{Optimal Completion Strategy Possibilistic Fuzzy C-Means (OCSPFCM) Algorithm}
The OCSPFCM algorithm is the first modification of the PFCM algorithm that we proposed for clustering incomplete data sets, the explanation is as follows.

\textbf{Step I}: Fix $m>1$, $\tau>1$, $\epsilon>0$ and $1<c<n$. Initiate $Y_{M}^{(0)}$, for each $y_{kj} \in Y_{M}$, with pick randomly available value in $Y_{C}$. Then pick $\textbf{v}^{(0)}\in\mathbb{R}^{s}$, $\textbf{v}^{(0)}$ can be chosen randomly from the $Y=\left \{\textbf{y}_{1},\textbf{y}_{2},\cdots,\textbf{y}_{n} \right\}\in\mathbb{R}^{p}$. Then at step $l$, $l=1,2,\cdots$

\textbf{Step II}: Calculate fuzzy membership degree ($u_{ik}$) which minimize the objective function $J_{m,\tau}$ using the following
\begin{equation}
u_{ik}^{(l)}=\left(\sum_{j=1}^{c}\left(\frac{d_{ik}^{2}}{d_{jk}^{2}}\right)^{\frac{1}{m-1}}\right)^{-1},1 \leq i \leq c ;~ 1 \leq k \leq n.
\end{equation}

\textbf{Step III}: Calculate possibilistic typical ($\delta_{i}$) which minimizes the objective function $J_{m,\tau}$ using the following

\begin{equation}
\delta_{i}^{(l)}=\frac{\sum_{k=1}^{n}\left(u_{ik}^{(l)}\right)^{m}d_{ik}^{2}}{\sum_{k=1}^{n}\left(u_{ik}^{(l)}\right)^{m}},1 \leq  k \leq n.
\end{equation}

\textbf{Step IV}: Calculate possibilistic membership degree ($t_{ik}$) which minimizes the objective function $J_{m,\tau}$ using the following

\begin{equation}
t_{ik}^{(l)}=\left(1+\left( \frac{\beta}{\delta_{i}}d_{ik}^{2}\right)^{\frac{1}{\tau-1}}\right)^{-1},1 \leq i \leq c ;~ 1 \leq k \leq n.
\end{equation}

\textbf{Step V}: Update cluster centre ($\textbf{v}_{i}$) which minimizes the objective function $J_{m,\tau}$ using the following

\begin{equation}
\textbf{v}_{i}^{(l)}=\frac{\sum_{k=1}^{n} \left(\left(\alpha u_{ik}^{(l)}\right)^{m} + \left(\beta t_{ik}^{(l)}\right)^{\tau}\right) \textbf{y}_{k}}{\sum_{k=1}^{n} \left(\left(\alpha u_{ik}^{(l)}\right)^{m} + \left(\beta t_{ik}^{(l)}\right)^{\tau}\right)}, 1 \leq i \leq c ;~1 \leq j \leq p.
\end{equation}

\textbf{Step VI}: Compare $\textbf{v}_{i}^{(l)}$ to $\textbf{v}_{i}^{(l-1)}$ using $\left\| \textbf{v}_{i}^{(l)}-\textbf{v}_{i}^{(l-1)}\right\|<\epsilon$. If true, then stop. Otherwise, go to \textbf{Step VII}.

\textbf{Step VII}: Update $Y_{M}$ which minimize the objective function $J_{m,\tau}$, for each $y_{kj} \in Y_{M}$ using the following

\begin{equation}
\textit{y}_{kj}^{(l)}=\frac{\sum_{k=1}^{n} \left(\left(\alpha u_{ik}^{(l)}\right)^{m} + \left(\beta t_{ik}^{(l)}\right)^{\tau}\right) \textit{v}_{ij}}{\sum_{k=1}^{n} \left(\left(\alpha u_{ik}^{(l)}\right)^{m} + \left(\beta t_{ik}^{(l)}\right)^{\tau}\right)}, 1 \leq i \leq c ;~1 \leq j \leq p.
\label{eq:10} 
\end{equation}
Now set $l=l+1$ and return to Step II.

Based on seven steps of the OCSPFCM algorithm described above, basicly \textbf{Step I} to \textbf{Step VI} are the PFCM algorithm. The difference is only on the process of missing values imputation which is done by initiating their with a random values that is available in the data sets (\textbf{Step I}). Then we update these values on \textbf{Step VII} by the sum of the fuzzy membership degrees and the possibilistic membership degrees multiplied by one of the values exist on the cluster center vector as shown in Equation \ref{eq:10}. 

\subsection{Nearest Prototype Strategy Possibilistic Fuzzy C-Means (NPSPFCM) Algorithm}

The NPSPFCM algorithm is the second modification of the PFCM algorithm that we proposed for clustering incomplete data sets. In general, the clustering process using the NPSPFCM algorithm similar to the OCSPFCM algorithm. The difference is only in \textbf{Step VII}, i.e. we update the missing values imputation by one value available on the nearest cluster center vector to itself. \textbf{Step VII} of the NPSPFCM algorithm is defined as follows.

\textbf{Step VII}: Update $Y_{M}^{(l)}$ which minimize the objective function $J_{m,\tau}$, for each $y_{kj} \in Y_{M}$ using the following

\begin{equation}
y_{kj}^{(l)}=v_{ij}^{(l)}, 1 \leq i \leq c; 1 \leq j \leq p,
\end{equation}
where $d_{ik}^{2}=\textbf{min} \left\{ d_{1k}^{2},d_{2k}^{2},...,d_{ck}^{2} \right\}$, $d_{1k}^{2} = \left \| \textbf{y}_{k} - \textbf{v}_{1}\right \|$, $d_{2k}^{2} = \left \| \textbf{y}_{k} - \textbf{v}_{2}\right \|$, and $d_{ck}^{2} = \left \| \textbf{y}_{k} - \textbf{v}_{c}\right \|$. Now set $l=l+1$ and return to \textbf{Step II}.

\section{Experimental Setup}\label{section:4}

In this study, we evaluated and demonstrated the potential of a modified PFCM algorithm for clustering incomplete data sets. The stages of the experiment are as follows. First, we cluster complete data sets using the PFCM algorithm to obtain the distribution of data points in the actual cluster. The result of this stage is used as a base in evaluating the performance of OCSPFCM and NPSPFCM algorithms. To obtain the optimal number of clusters in a complete data set, we use a cluster validity index. The cluster validity index used is the Xie-Beni index shown in Equation (\ref{eq:12}). A validity index is a measure used to determine the optimal number of clusters. We used the Xie-Beni index by reason of the \textit{partition coefficient} (PC) and \textit{classification entropy} (CE) index ignore cluster center and data in the index calculation. Whereas the cluster center and data are two basic things involved in the process of clustering based on fuzzy clustering rule \cite{rustam1}. The optimal number of clusters are indicated by the smallest Xie-Beni index value. Xie and Beni \cite{xie} propose the cluster validity index as follows.

\begin{equation}
\textbf{XB}(U,V;X) = \frac{\sum_{i=1}^{c}\sum_{k=1}^{n}(u_{ik})^{m}\left\Vert \textbf{x}_{k} - \textbf{v}_{i} \right\Vert}{n\cdot \underset{i\neq j}{\textbf{min}}(\textbf{v}_{i} - \textbf{v}_{j})}
\label{eq:12}
\end{equation}

\begin{figure}[h!]
	\centering
	\includegraphics[width=8cm,height=6cm]{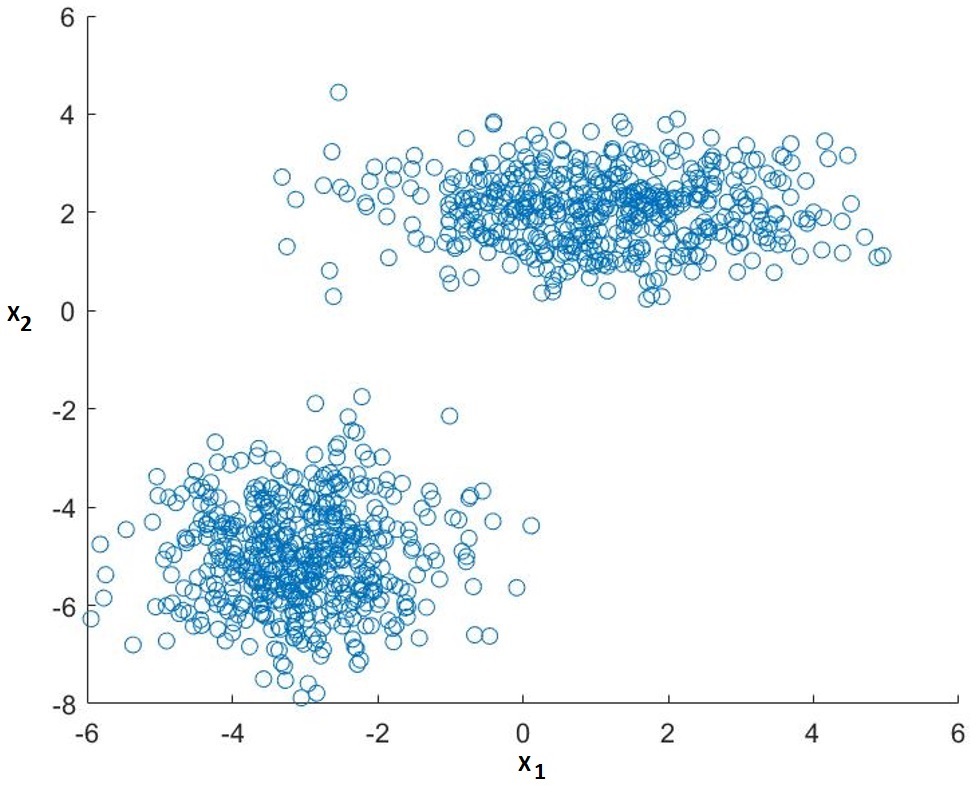} 
	\caption{Artificial data sets I}
	\label{fig:0}
\end{figure}

The second stage taken is the complete data sets are made contain the missing values. In this study, the performance of OCSPFCM and NPSPFCM algorithms is evaluated on real world and artificial data sets. The real world data sets used is iris \cite{fisher} and wine \cite{forina} data sets downloaded from \url{http://archive.ics.uci.edu/ml} \cite{Dua}. Iris data sets consist of 150 data points with 4 features per data point. In matrix form, iris data sets have a size of [150$\times$4]. Wine data sets consist of 178 data points with 13 features per data point. In matrix form, wine data sets have a size [178$\times$13]. While the artificial data sets used is the artificial data sets generated using the Gaussian mixture distribution rule. We generate artificial data sets with two components. Artificial data sets I consists of 1000 data points with 2 features per data point. In matrix form, artificial data sets I have a size [1000$\times$2]. A scatter plot of artificial data set I shown in Figure \ref{fig:0}. Artificial data sets II consists of 1000 data points with 14 features per data point. In matrix form, artificial data set II have a size [1000$\times$14]. The rows and columns of the data sets matrix represent the number of data points and features, respectively. Complete iris, wine, artificial data set I and II are made contain missing values with predetermined percentages. The percentage of the number of missing values for each data sets is 5\%, 10\%, 15\%, 20\%, 25\%, and 30\%, respectively. We make complete data sets contains missing values by a random way. The value to be used as the missing values are chosen randomly in the column vector direction (feature) of matrix data sets.

The third stage is complete data sets that have been become incomplete data sets is clustered using OCSPFCM and NPSPFCM algorithms. The performance of OCSPFCM and NPSPFCM algorithms in clustering incomplete data sets are evaluated in three ways: the percentage accuracy of each data point being on the correct cluster member, number of iterations to termination, and centroid errors. The formula used to calculate the percentage of accuracy is as follows  \cite{rustam}.

\begin{equation}
\% \: accuracy = \frac{a}{n}100\%
\label{eq:13}
\end{equation}
Where $a$ is the number of data points clustered correctly, and $n$ is the total number of data points. In this study, the centroid errors are the magnitude of the error of cluster center for incomplete data sets clustered using OCSPFCM and NPSPFCM algorithms when compared to the cluster center of complete data sets clustered using the PFCM algorithm. In some applications, information about the cluster center is important to know the data point partitioning in the cluster \cite{himmelspach}. Therefore, we also evaluate the two algorithms by calculating the centroid errors at each level of the missing values percentage. Euclid distance formula is used to calculate the centroid errors. Then this centroid errors ($ce$) are averaged using the following formula:

\begin{equation}
ce = \frac{\sum_{i=1}^{c} ce_{i}}{c}.
\end{equation}
Where $ce_{i}$ is the $i$-th centroid error. 

\section{Experimental Results and Discussions} \label{section:5}

In accordance with the experimental setup, the first thing to do is to cluster the complete data sets using the PFCM algorithm with the Xie-Beni index as cluster validity index. The result obtained indicate that the smallest Xie-Beni index value for each data sets is in the two clusters. This means that the optimal number of clusters to the complete iris data sets is two clusters. This result in line with the results obtained by Pakhira et al. \cite{pakhira} with the Davies-Bouldin (DB) index and the Dunn's index \cite{davies} as the cluster validity index. Similarly, for the complete wine data sets, the optimal number of clusters obtained with the two clusters. This result is in line with the results obtained by Zhang et al. \cite{zhang} with the MPC index and the MPA index \cite{dave} as cluster validity index. For artificial data sets I and II, the optimal number of clusters obtained in the two clusters as well. This corresponds to the data sets of two components generated using Gaussian mixture distribution. Details of those cluster results shown in Table \ref{tab:tabel1}. 

\begin{table}[h!]
	\centering
	\caption{The number of data points for each cluster on complete data sets}
	\begin{tabular}{ccc}
		\hline
		\textbf{} &  \multicolumn{2}{c}{\textbf{\textbf{ Cluster}}} \\
		\cline{2-3}
		\textbf{} & \textbf{I} & \textbf{II}  \\
		\hline
		\textbf{Iris} & 50 data points & 100 data points  \\
		\hline
		\textbf{Wine} & 78 data points & 100 data points  \\
		\hline
		\textbf{Artificial I} & 494 data points & 506 data points  \\
		\hline
		\textbf{Artificial II} & 510 data points & 490 data points  \\
		\hline
	\end{tabular}
	\label{tab:tabel1}
\end{table}

Table \ref{tab:tabel1} shows for complete iris data sets, 50 data points to be members of the first cluster, and 100 data points become members of the second cluster. For complete wine data sets, 78 data points become members of the first cluster, and 100 data points become members of the second cluster. For complete artificial data sets I, 494 data points become members of the first cluster and 506 data points become members of the second cluster. While, for complete artificial data sets II, 510 data points become members of the first cluster and 490 data points become members of the second cluster. These results will be a base to evaluate the performance of the modified PFCM algorithm that we are proposing. Due to differences in the percentage of accuracy, number of iterations to termination, and centroid errors in every experiment, and to gain a representative result, we conducted 30 experiments for each thing that evaluated. So that each of the three things that are evaluated will have 30 values. The average of the 30 values is the value that will be used to represent the percentage of accuracy, number of iterations to termination, and centroid errors, respectively.

\subsection{Experiment on Iris Data Sets} \label{subsection:iris}

The results of the complete iris data sets cluster by the PFCM algorithm is used as a base to evaluate the performance of the OCSPFCM and NPSPFCM algorithms on iris data sets experiment. The first evaluation is the percentage of accuracy.

\begin{figure}[h!]
	\centering
	\includegraphics[width=8cm,height=6cm]{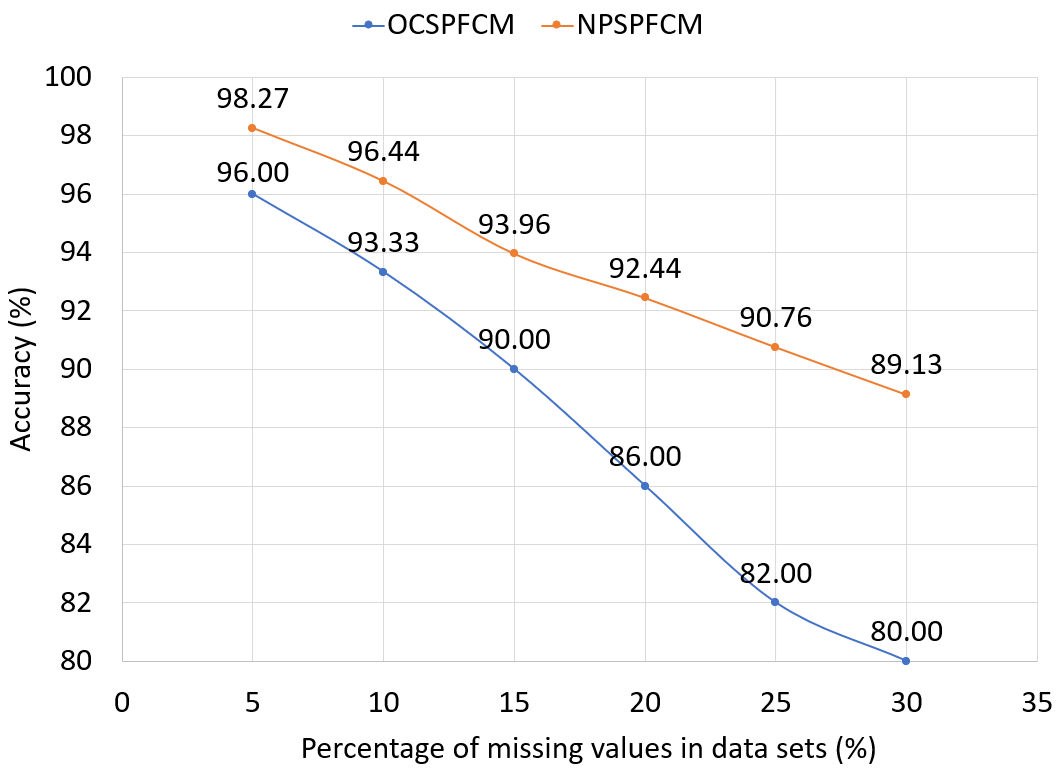} 
	\caption{The average accuracy percentage on iris data sets}
	\label{fig:1}
\end{figure}

Figure \ref{fig:1} shows the average accuracy percentage for iris data sets using OCSPFCM and NPSPFCM algorithms. For missing values of below 15\%, the OCSPFCM algorithm is an accuracy percentage above 90\%. However, at the number of missing values between 20\% to 30\%, the OCSPFCM algorithm only obtained an accuracy percentage of above 80\% and a maximum is 86\%. The NPSPFCM algorithm is an accuracy percentage above 90\% for all the missing values tested, expect for the 30\% missing values with an accuracy of 89.13\%. The percentage of accuracy shows a significant difference at above 20\% of the total missing value. Figure \ref{fig:1} also shows that the greater the number of missing values, the lower the percentage accuracy. The decrease in the percentage of accuracy is caused by update the missing value imputation falls far from the value that should be. So that the data points that contain the missing values previously being a member of the inappropriate cluster. 80\% of accuracy percentage on the OCSPFCM algorithm means that with a 30\% missing value, there are 130 data points out of a total of 150 data points being members of the right cluster. Instead, there are 20 data points that be members of the inappropriate cluster. While 89.13\% of accuracy percentage on the NPSPFCM algorithm means 134 data points are members of the right cluster. In contrast, there are 16 data points that be members of the inappropriate cluster. 

\begin{figure}[h!]
	\centering
	\includegraphics[width=8cm,height=6cm]{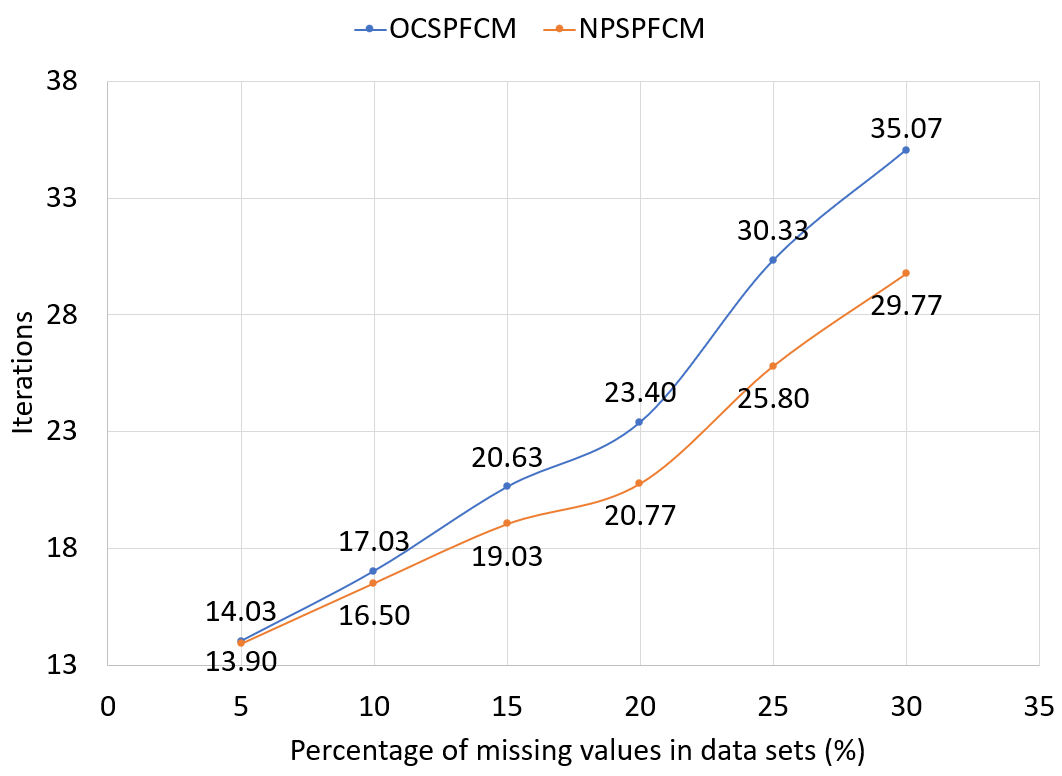} 
	\caption{The average number of iterations to termination on iris data sets}
	\label{fig:2}
\end{figure}

Figure \ref{fig:2} shows the number of iterations to termination needed by OCSPFCM and NPSPFCM algorithms on the iris data sets. Figures \ref{fig:1} and \ref{fig:2} show the behavior that is inversely proportional to the percentage of accuracy and the number of iterations. The percentage of accuracy decreases inversely with an increase in the number of iterations needed to terminate. An increase in the number of missing values caused an increase in the number of iterations as well. In other words, the greater the number of missing values, the more iterations needed to termination.

\begin{figure}[h!]
	\centering
	\includegraphics[width=8cm,height=6cm]{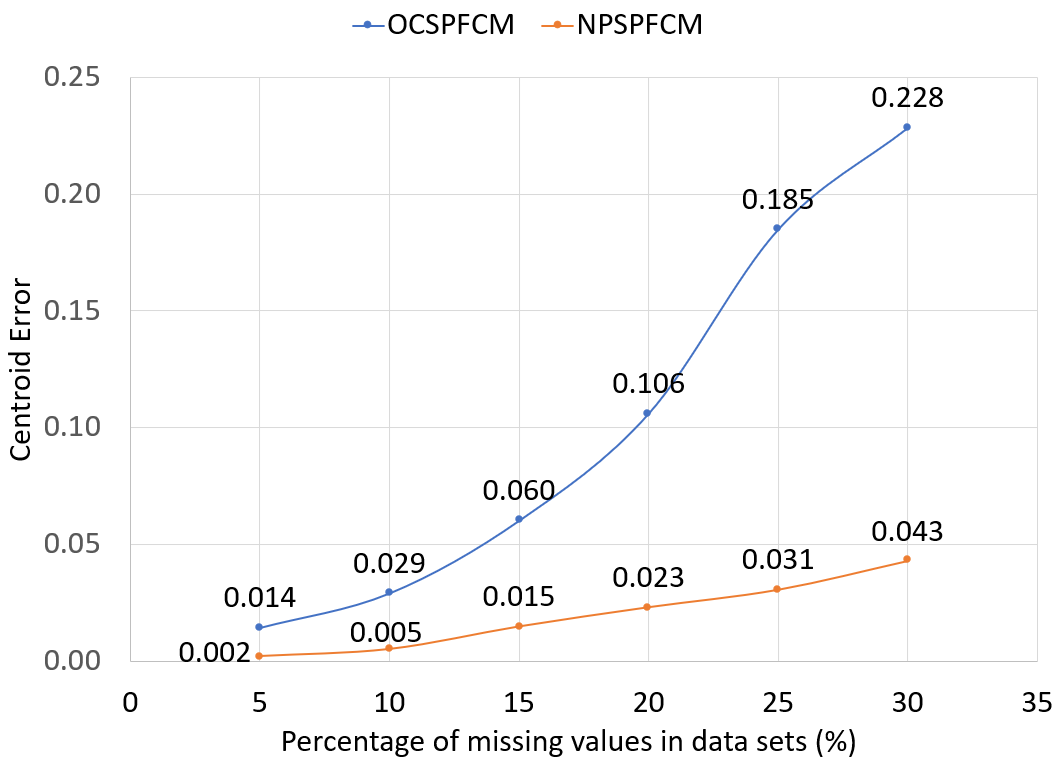} 
	\caption{The average centroid errors on iris data sets}
	\label{fig:3}
\end{figure}

Figure \ref{fig:3} shows the average centroid errors for iris data sets using OCSPFCM and NPSPFCM algorithms. We can see that the difference in centroid errors between the OCSPFCM algorithm and NPSPFCM begins significantly at the 20\% missing values. The shift of the cluster center is closely related to the process of updating the missing values imputation. The algorithm updates the missing value imputation falls far from the actual value, which occurrence a cluster center error. Figure \ref{fig:3} also shows that the process of updating the missing values imputation by the NPSPFCM algorithm has smaller cluster center (centroid) errors compared to the OCSPFCM algorithm.  

\subsection{Experiment on Wine Data Sets} \label{subsection:wine}

\begin{figure}[h!]
	\centering
	\includegraphics[width=8cm,height=6cm]{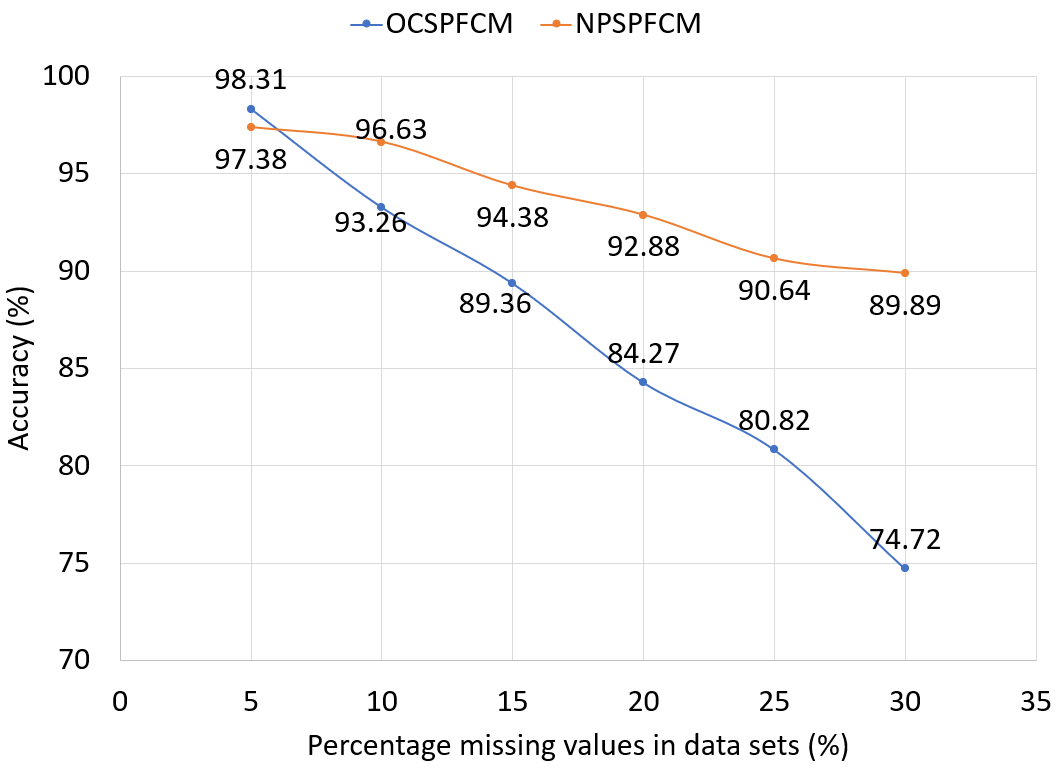} 
	\caption{The average accuracy percentage on wine data sets}
	\label{fig:4}
\end{figure}

\begin{figure}[h!]
	\centering
	\includegraphics[width=8cm,height=6cm]{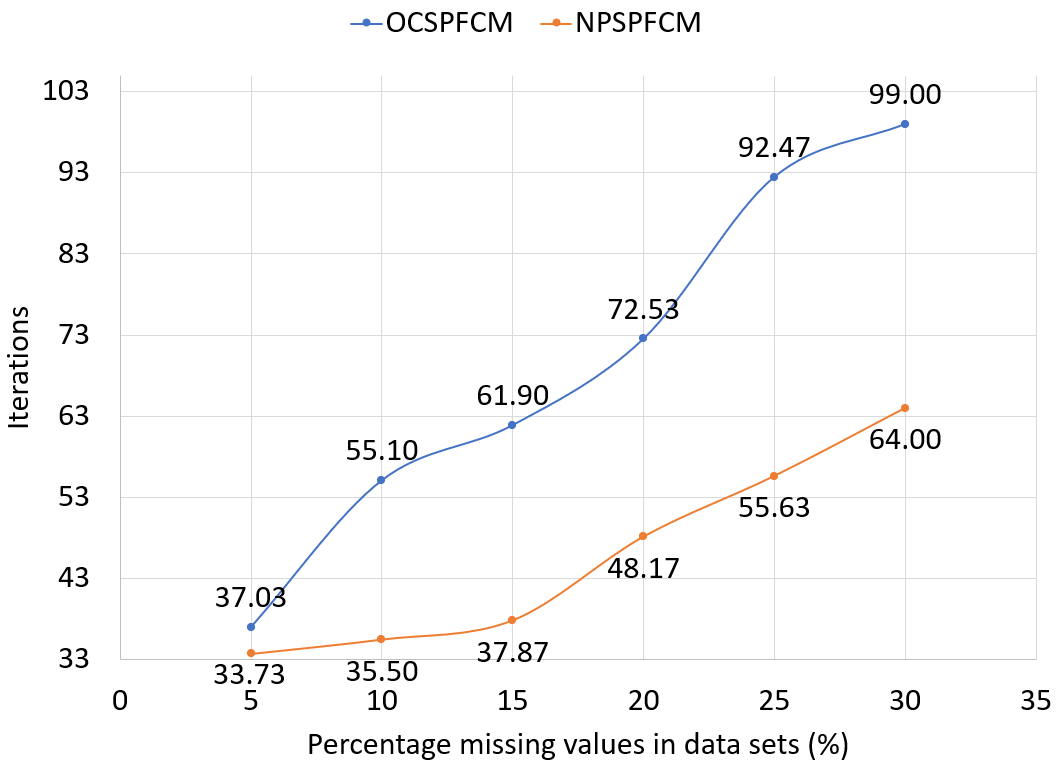} 
	\caption{The average number of iterations to termination on wine data sets}
	\label{fig:5}
\end{figure}

\begin{figure}[h!]
	\centering
	\includegraphics[width=8cm,height=6cm]{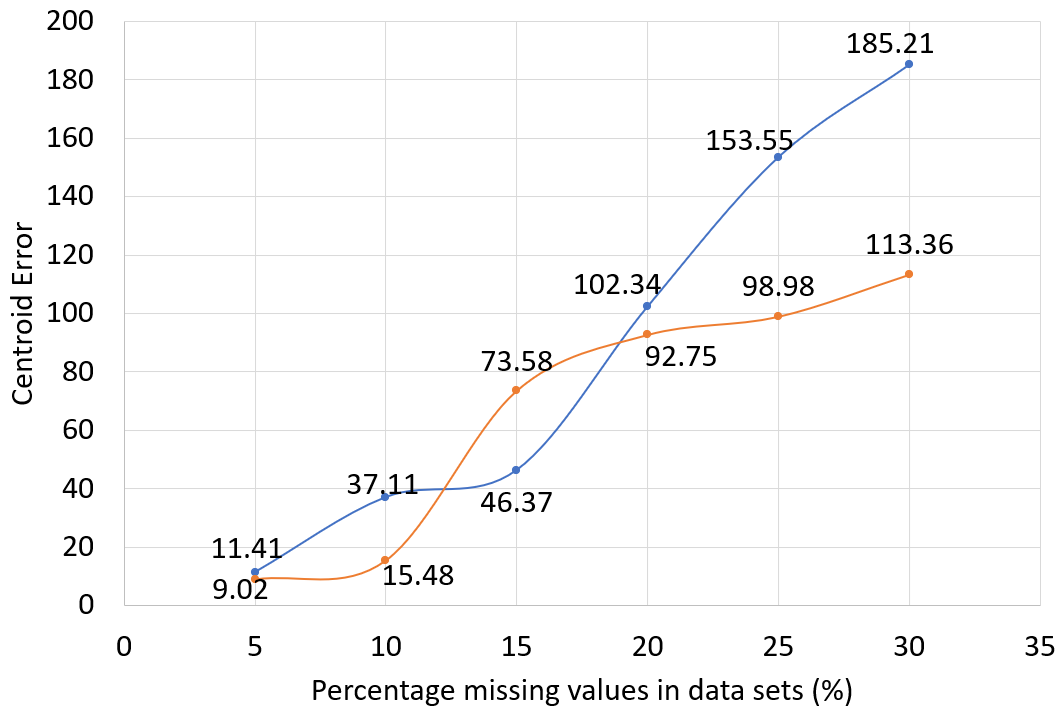} 
	\caption{The average centroid errors on wine data sets}
	\label{fig:6}
\end{figure}

In the wine data set, evaluations related to the percentage accuracy, the number of iterations, and centroid errors of the OCSPFCM and NPSPFCM algorithms are respectively shown in Figures \ref{fig:4}, \ref{fig:5}, and \ref{fig:6}.

Figure \ref{fig:4} shows the average of accuracy percentage of the OCSPFCM and NPSPFCM algorithms in the wine data sets. The OCSPFCM algorithm shows the percentage of accuracy above 90\% (for the number of missing values 5\% and 10\%), above 80\% (for the number of missing values 15\%, 20\%, and 25\%), and 74.72\% (for 30\% missing value). Whereas the NPSPFCM algorithm shows an accuracy percentage above 90\% for all levels of missing values, except for a 30\% level of missing value with an accuracy of 89.89\%. Both those algorithms produced a percentage of accuracy that decreases along with the greater percentage of missing value. In the OCSPFCM algorithm, the percentage of accuracy is 74.72\% means that with a 30\% missing values there are 133 data points out of a total of 178 data points being members of the right cluster. Conversely, there are 45 data points that being members of the inappropriate cluster. In the NPSPFCM algorithm, the percentage of accuracy is 89.89\%  means that there are 160 data points out of a total of 178 data points being members of the correct cluster at the 30\% missing value level. In contrast, there are 18 data points that are members of the inappropriate cluster.

Figure \ref{fig:5} shows the average number of iterations to termination on wine data sets using the OCSPFCM and NPSPFCM algorithms. In general, the number of iterations of the two algorithms increases with the upward in the number of missing values. The NPSPFCM algorithm provided a number of more efficient iterations than the OCSPFCM algorithm to termination.  

Figure \ref{fig:6} shows the average centroid errors in wine data sets using the OCSPFCM and NPSPFCM algorithms. Based on figure \ref{fig:6}, we can see a greater centroid error on the OCSPFCM algorithm than the NPSPFCM algorithm at all levels of the total missing values, except at the level of 15\% the missing values, NPSPFCM gives a greater centroid errors.

\subsection{Experiment on Artificial Data Sets I} \label{subsection:artificial1}

\begin{figure}[h!]
	\centering
	\includegraphics[width=8cm,height=6cm]{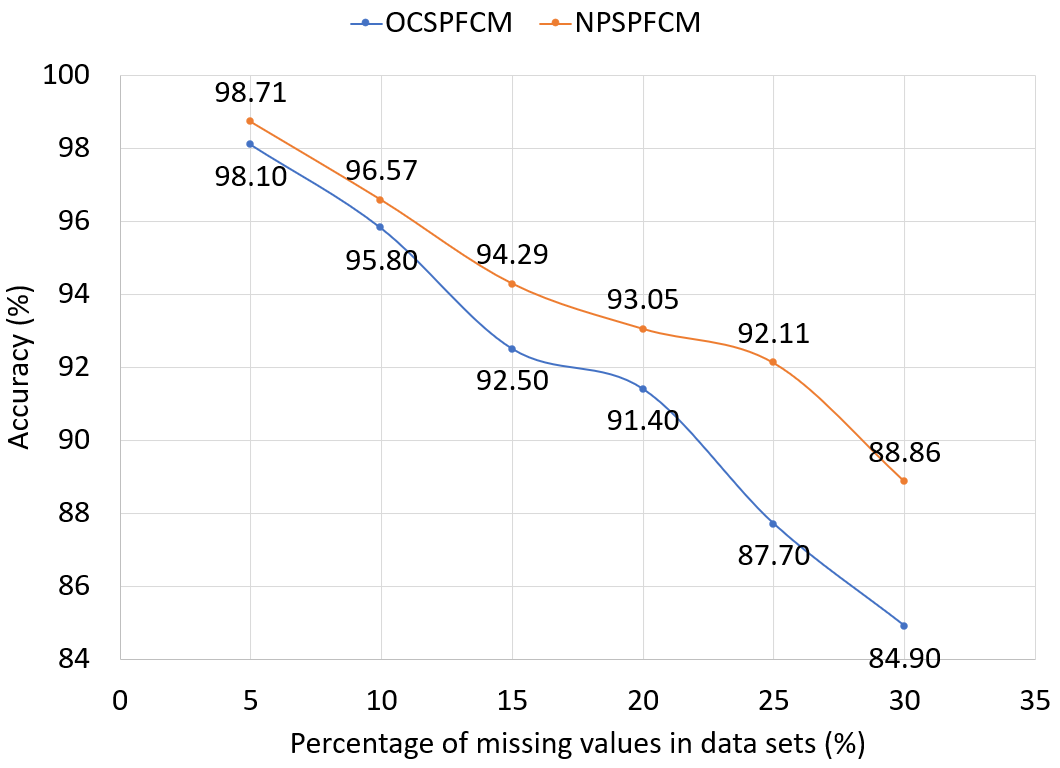} 
	\caption{The average accuracy percentage on artificial data sets I}
	\label{fig:7}
\end{figure}

\begin{figure}[h!]
	\centering
	\includegraphics[width=8cm,height=6cm]{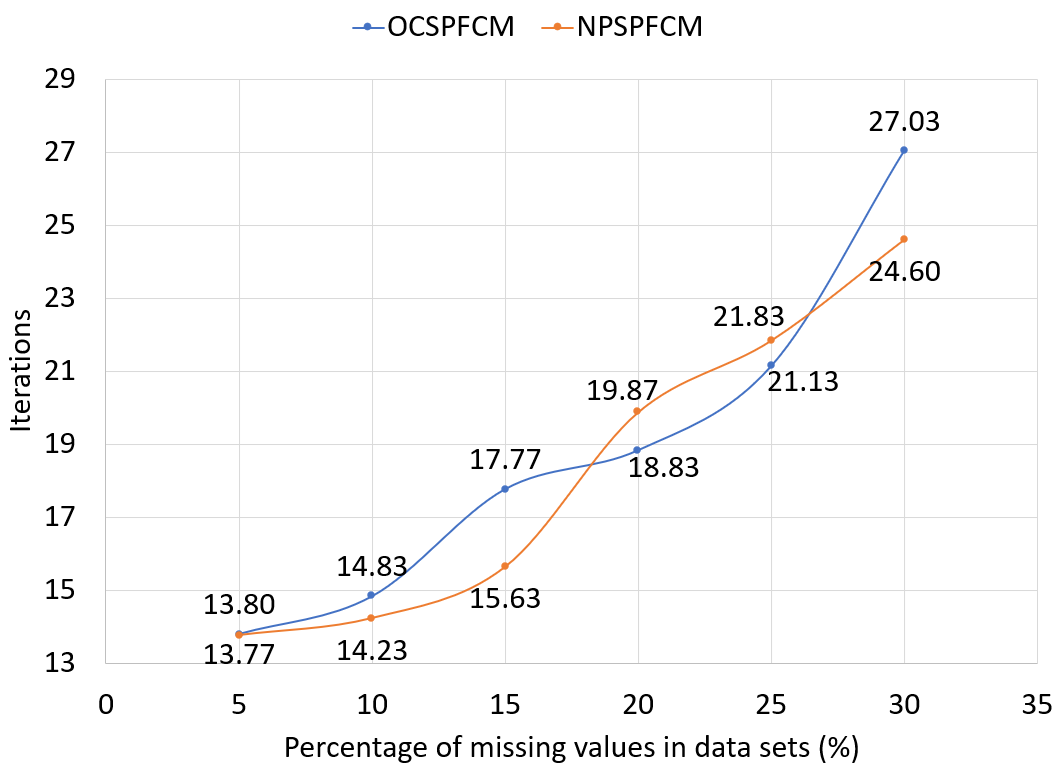} 
	\caption{The average number of iterations to termination on artificial data sets I}
	\label{fig:8}
\end{figure}

\begin{figure}[h!]
	\centering
	\includegraphics[width=8cm,height=6cm]{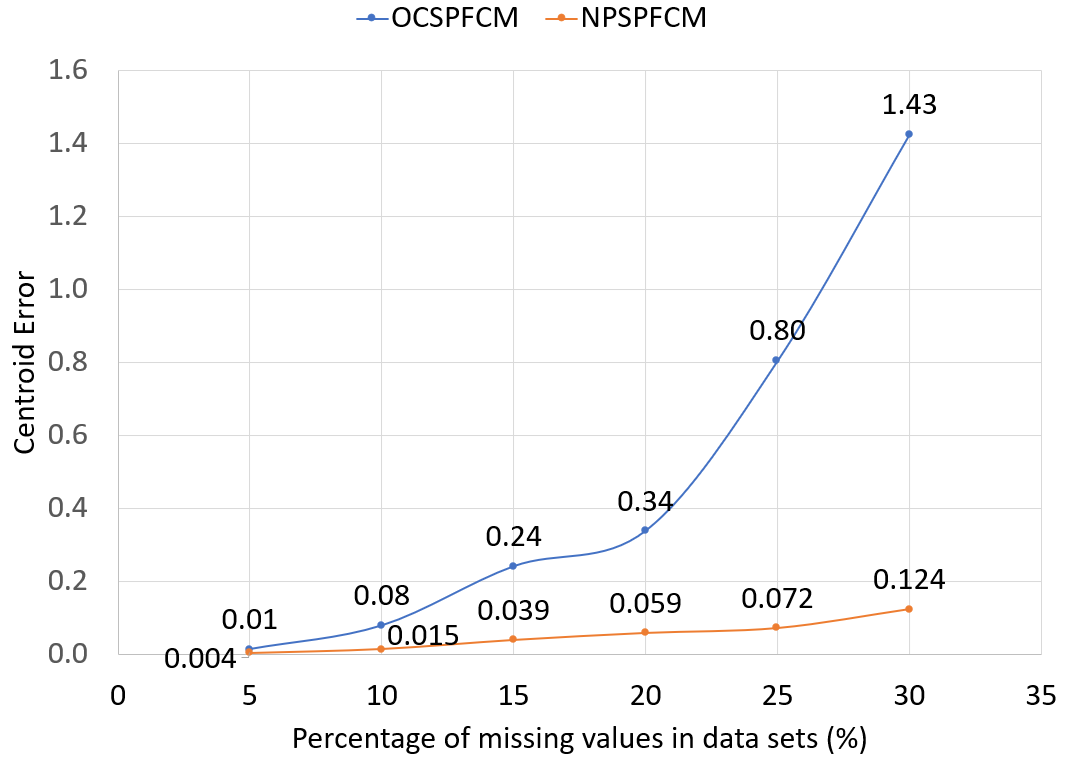} 
	\caption{The average centroid errors on artificial data sets I}
	\label{fig:9}
\end{figure}

In the artificial data set I, evaluations related to the percentage of accuracy, number of iterations, and centroid error of the OCSPFCM and NPSPFCM algorithms are respectively shown in Figures \ref{fig:7}, \ref{fig:8}, and \ref{fig:9}. 

Figure \ref{fig:7} shows the accuracy percentage of the OCSPFCM and NPSPFCM algorithms on the artificial data sets I. The OCSPFCM algorithm below 20\% of the missing values gives an accuracy percentage above 90\%. This result is higher than what obtained in the iris data sets, where the OCSPFCM algorithm gives an accuracy percentage above 90\% at 15\% missing value. As for the NPSPFCM algorithm, the performance is similar in the iris data sets for all missing value levels, the NPSPFCM algorithm gives an accuracy percentage above 90\%, except for 30\% missing values, with accuracy percentage is 88.86\%. This 88.86\% accuracy percentage value on the NPSPFCM algorithm means that with 30\% missing values there are 886 data points out of a total of 1000 data points that are members of the right cluster. Instead, there are 114 data points that are members of the cluster that is not right. While on the OCSPFCM algorithm, 84.90\% means that there are 849 data points out of a total of 1000 data points being members of the right cluster. In contrast, there are 151 data points that are members of the inappropriate cluster. In general, both algorithms have a percentage of accuracy that decreases with an increasing number of missing values.

Figure \ref{fig:8} shows the average number of iterations required by OCSPFCM and NPSPFCM algorithms to termination. Although there is a change in the number of iterations between the two algorithms. In general, the number of iterations required by the two algorithms to terminate relatively similar at each level of the missing value.

Figure \ref{fig:9} shows the average centroid errors of the NPSPFCM algorithm which is smaller than the OCSPFCM algorithm. This is the implication of the process of updating the imputation of missing values by the NPSPFCM algorithm which gives a smaller cluster center (centroid) error than the OCSPFCM algorithm. In other words, the cluster center generated by the NPSPFCM algorithm falls closer to the base centroid used in the complete artificial data sets I.

\subsection{Experiment on Artificial Data Sets II} \label{subsection:artificial2}

\begin{figure}[h!]
	\centering
	\includegraphics[width=8cm,height=6cm]{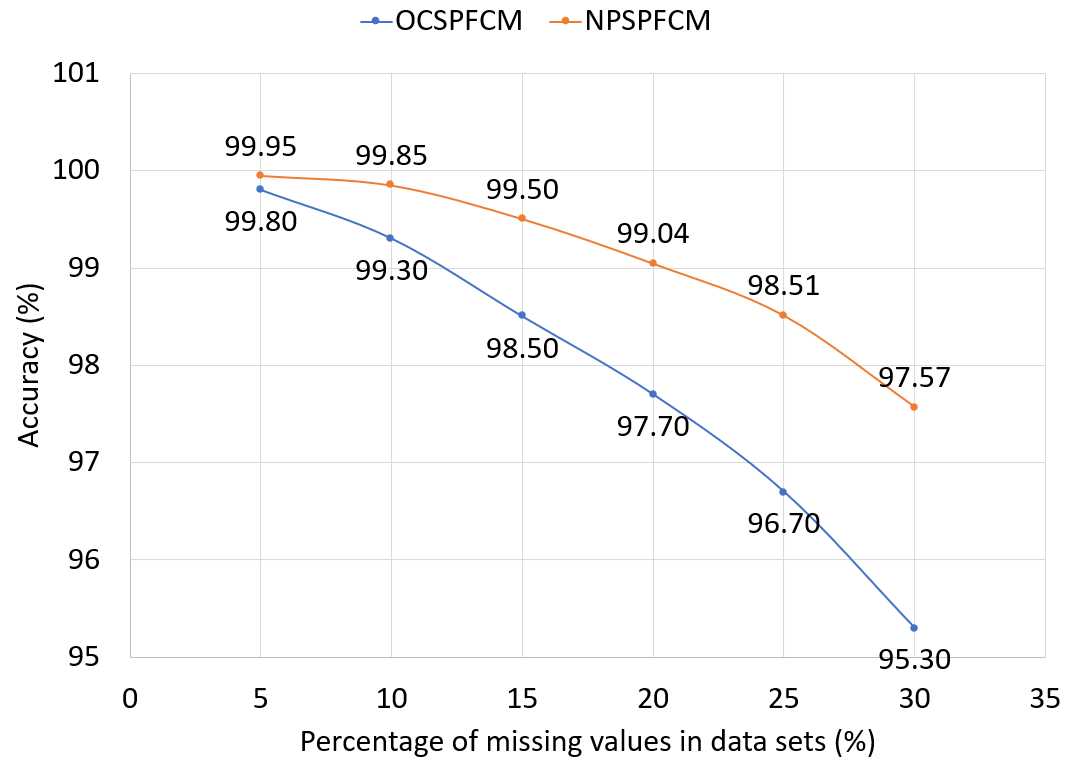} 
	\caption{The average accuracy percentage on artificial data sets II}
	\label{fig:10}
\end{figure}

\begin{figure}[h!]
	\centering
	\includegraphics[width=8cm,height=6cm]{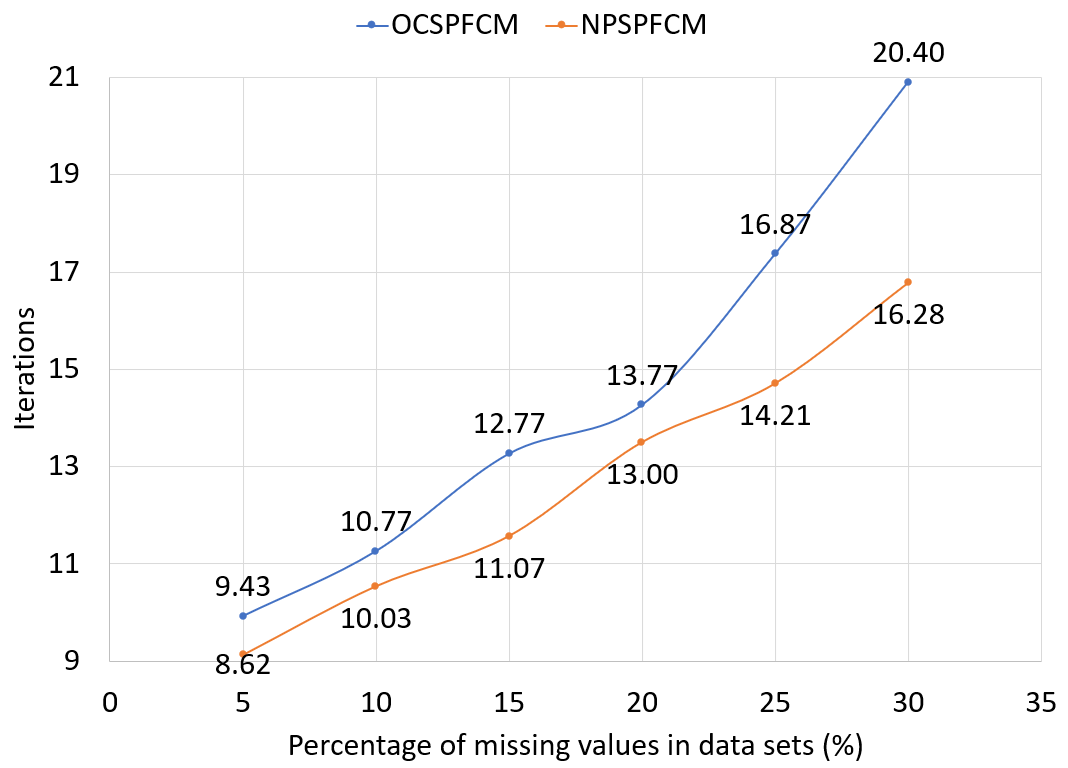} 
	\caption{The average number of iterations to termination on artificial data sets II}
	\label{fig:11}
\end{figure}

\begin{figure}[h!]
	\centering
	\includegraphics[width=8cm,height=6cm]{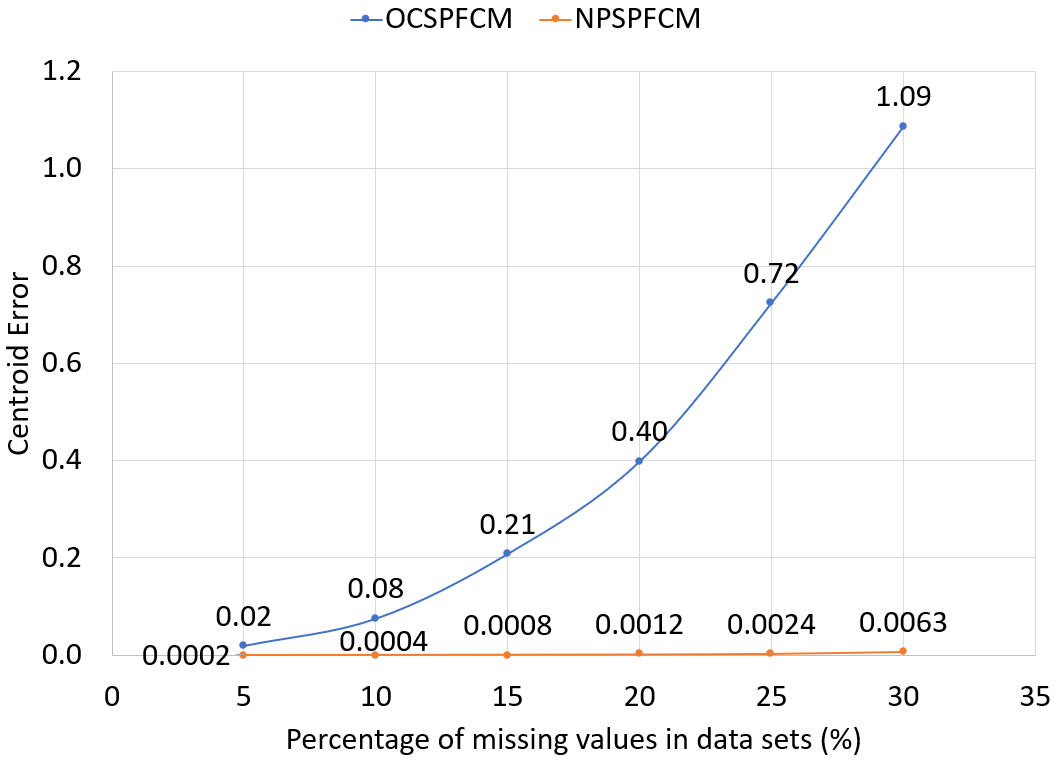} 
	\caption{The average centroid errors on artificial data sets II}
	\label{fig:12}
\end{figure}

In the artificial data sets II, evaluations related to the percentage of accuracy, the number of iterations, and the centroid errors of the OCSPFCM and NPSPFCM algorithms are respectively shown in Figures \ref{fig:10}, \ref{fig:11}, and \ref{fig:12}.

Figure \ref{fig:10} shows the average of accuracy percentage in the OCSPFCM and NPSPFCM algorithms in the artificial data sets II. Both algorithms provide accuracy percentages above 95\% for all missing values percentage levels, which is an accurate value. For the 30\% missing value, the OCSPFCM and NPSPFCM algorithms show an accuracy percentage of 95.30\% and 97.57\%, respectively. That means for the OCSPFCM algorithm there are available 953 data points out of a total of 1000 data points that members of the right cluster. Instead, there are 47 data points that members of an inappropriate cluster. Whereas in the NPSPFCM algorithm there are 975 data points that members of the right cluster. Instead, there are 25 data points that members of an inappropriate cluster.

Figure \ref{fig:11} shows the number of iterations is more efficient provided by the NPSPFCM algorithm. In addition to more efficient iterations, the NPSPFCM algorithm also provides a higher percentage of accuracy as shown in Figure \ref{fig:10}. It was also inversely shown in Figures \ref{fig:10} and \ref{fig:11}, which is the percentage of accuracy that decreases and the number of iterations increases with an increasing number of missing values.  

In addition to the higher percentage of accuracy and the number of iterations that more efficient, the NPSPFCM algorithm also produced smaller centroid errors compared to the OCSPFCM algorithm as shown in Figure \ref{fig:12}.  

The OCSPFCM and NPSPFCM algorithms provide accuracy percentages significantly different for the number of features almost the same, but significantly different in the number of data points as shown in Figure \ref{fig:4} and \ref{fig:10}. However, we also find different results result in the iris data sets experiment shown in Figure \ref{fig:1} and the experiment in the artificial data sets I as shown in Figure \ref{fig:7}, that the addition of the number of data points does not significantly affect the acquisition of the percentage of accuracy. Therefore, we can conclude that the OCSPFCM and NPSPFCM algorithms provide a more accurate percentage of accuracy in data sets with a greater number of data points and features. While data sets with a smaller number of features, although the addition of data points becomes larger, the two algorithms do not provide a significant increase in the percentage of accuracy.

The performance of the NPSPFCM algorithm on the number of iterations to termination in all data sets is also always more efficient than the OCSPFCM algorithm, except for the artificial data sets I whose number of iterations was alternately larger. But in general, the number of iterations to termination in the artificial data sets I give a relatively equal number of iterations between the OCSPFCM and NPSPFCM algorithms.The excellence of the NPSPFCM algorithm in the efficiency of the number of iterations is due to the NPSPFCM algorithm updating the imputation of the missing values using one of the values available at the cluster center vector nearest to itself. This accelerates the convergence of cluster centers directly, or $\left\|\textbf{v}_{i}^{(l)}-\textbf{v}_{i}^{(l-1)}\right\|<\epsilon$ faster achieved on the NPSPFCM algorithm than OCSPFCM algorithm. Whereas in the OCSPFCM algorithm, the imputation of the missing values updated using the sum of the degrees of fuzzy membership with the possibilistic of membership degrees then multiplied by one of the values available in the existing cluster center vector. From this, we understood that the value used to update the missing value may be one of the values of the cluster center which is not the cluster center where the imputation of the missing value is. So this causes the slow convergence of cluster centers.

Finally, we have evaluated the performance of OCSPFCM and NPSPFCM algorithms on centroid errors in each data sets. The results obtained the centroid errors of the NPSPFCM algorithm on all data sets are always smaller than the OCSPFCM algorithm, except for the wine data sets. However, generally the NPSPFCM algorithm always gives a smaller centroid errors in the wine data sets. The smaller centroid errors in the NPSPFCM algorithm can be interpreted as the ability of the NPSPFCM algorithm to produce cluster centers on incomplete data set with a 'location' not far from cluster centers of the complete data set. This is the implication of the process of updating the imputation of the missing values, where the NPSPFCM algorithm produces value not far from the value that should be. This is also the reason why the NPSPFCM algorithm obtains a more accurate percentage of accuracy and the number of iterations to termination more efficiently.

\label{sect:Pha}

\section{Conclusions}\label{section:6}

In this paper, we have presented the potential and performance modification of the PFCM algorithm for clustering incomplete data sets. The modification of the PFCM algorithm, we call as OCSPFCM and NPSPFCM algorithms. This paper divided into three stages. In the first stage, we conducted a clustering of complete data sets using the PFCM algorithm. The cluster results obtained at this stage become a base in evaluating the performance of the OSCPFCM and NPSPFCM algorithms. We evaluated the performance of the two algorithms on three ways: accuracy percentage, a number of iterations to termination, and centroid errors. In the second stage, the complete data sets were made into an incomplete data set, in this case, the data set contains the missing values with the predetermined percentage. In the third stage, we clustered incomplete data sets using the OCSPFCM and NPSPFCM algorithms. The results showed both algorithms have the potential to clustering incomplete data sets. However, the NPSPFCM algorithm has better performance than the OCSPFCM algorithm based on three things that are evaluated. Furthermore, the modification of the PFCM algorithm proposed in this paper can provide knowledge in the algorithm for incomplete data sets clustering.


\bibliographystyle{unsrt}

\bibliography{references}

\begin{thebibliography}{10}

\bibitem{himmelspach}
Ludmila Himmelspach.
\newblock {\em Fuzzy clustering of incomplete data}.
\newblock PhD thesis, 2016.

\bibitem{bezdek1}
James~C Bezdek, Robert Ehrlich, and William Full.
\newblock Fcm: The fuzzy c-means clustering algorithm.
\newblock {\em Computers \& Geosciences}, 10(2-3):191--203, 1984.

\bibitem{krishnapuram}
Raghuram Krishnapuram and James~M Keller.
\newblock A possibilistic approach to clustering.
\newblock {\em IEEE transactions on fuzzy systems}, 1(2):98--110, 1993.

\bibitem{bezdek2}
Richard~J Hathaway and James~C Bezdek.
\newblock Fuzzy c-means clustering of incomplete data.
\newblock {\em IEEE Transactions on Systems, Man, and Cybernetics, Part B
  (Cybernetics)}, 31(5):735--744, 2001.

\bibitem{dixon}
John~K Dixon.
\newblock Pattern recognition with partly missing data.
\newblock {\em IEEE Transactions on Systems, Man, and Cybernetics},
  9(10):617--621, 1979.

\bibitem{bezdek3}
Nikhil~R Pal, Kuhu Pal, James~M Keller, and James~C Bezdek.
\newblock A possibilistic fuzzy c-means clustering algorithm.
\newblock {\em IEEE transactions on fuzzy systems}, 13(4):517--530, 2005.

\bibitem{rustam1}
Rustam, Agus~Y Gunawan, and Made Tri Ari~P Kresnowati.
\newblock The hard c-means algorithm for clustering indonesian plantation
  commodity based on metabolites composition.
\newblock In {\em Journal of Physics: Conference Series}, volume 1315, page
  012085. IOP Publishing, 2019.

\bibitem{xie}
Xuanli~Lisa Xie and Gerardo Beni.
\newblock A validity measure for fuzzy clustering.
\newblock {\em IEEE Transactions on Pattern Analysis \& Machine Intelligence},
  (8):841--847, 1991.

\bibitem{fisher}
Ronald~A Fisher.
\newblock The use of multiple measurements in taxonomic problems.
\newblock {\em Annals of eugenics}, 7(2):179--188, 1936.

\bibitem{forina}
M~Forina, S~Lanteri, C~Armanino, et~al.
\newblock Parvus-an extendible package for data exploration, classification and
  correlation, institute of pharmaceutical and food analysis and technologies,
  via brigata salerno, 16147 genoa, italy (1988).
\newblock {\em Av. Loss Av. O set Av. Hit-Rate}, 1991.

\bibitem{Dua}
Dheeru Dua and Casey Graff.
\newblock {UCI} machine learning repository.
\newblock 2017.
\newblock \url{http://archive.ics.uci.edu/ml}.

\bibitem{rustam}
Rustam, A~Yodi Gunawan, and M~T A~Penia Kresnowati.
\newblock Artificial neural network approach for the identification of clove
  buds origin based on metabolites composition.

\bibitem{pakhira}
Malay~K Pakhira, Sanghamitra Bandyopadhyay, and Ujjwal Maulik.
\newblock Validity index for crisp and fuzzy clusters.
\newblock {\em Pattern recognition}, 37(3):487--501, 2004.

\bibitem{davies}
David~L Davies and Donald~W Bouldin.
\newblock A cluster separation measure.
\newblock {\em IEEE transactions on pattern analysis and machine intelligence},
  (2):224--227, 1979.

\bibitem{zhang}
Dawei Zhang, Min Ji, Jun Yang, Yong Zhang, and Fuding Xie.
\newblock A novel cluster validity index for fuzzy clustering based on
  bipartite modularity.
\newblock {\em Fuzzy Sets and Systems}, 253:122--137, 2014.

\bibitem{dave}
Rajesh~N Dave.
\newblock Validating fuzzy partitions obtained through c-shells clustering.
\newblock {\em Pattern recognition letters}, 17(6):613--623, 1996.

\end{thebibliography}

\end{document}